\documentclass{article}

\usepackage{PRIMEarxiv}

\usepackage[utf8]{inputenc} 
\usepackage[T1]{fontenc}    
\usepackage{hyperref}       
\usepackage{url}            
\usepackage{booktabs}       
\usepackage{amsfonts}       
\usepackage{nicefrac}       
\usepackage{microtype}      
\usepackage{lipsum}
\usepackage{fancyhdr}       
\usepackage{graphicx}       
\usepackage{amsmath}
\usepackage{booktabs}
\usepackage{microtype}
\usepackage{multirow}
\usepackage{natbib}
\usepackage{fancyvrb}
\usepackage{endnotes}
\usepackage{bigfoot}
\usepackage{etoolbox}

\AtBeginEnvironment{quotation}{\noindent\ignorespaces}

\title{A Content-Based Novelty Measure for Scholarly Publications: A Proof of Concept\thanks{This research was supported in part by Lilly Endowment, Inc., through its support for the Indiana University Pervasive Technology Institute.}}

\author{
  Haining Wang \\
  Indiana University Bloomington\\
  Bloomington, Indiana, USA \\
  \texttt{hw56@indiana.edu} \\}

\begin{document}
\maketitle

\begin{abstract}
Novelty, akin to gene mutation in evolution, opens possibilities for scholarly advancement. Although peer review remains the gold standard for evaluating novelty in scholarly communication and resource allocation, the vast volume of submissions necessitates an automated measure of scholarly novelty. Adopting a perspective that views novelty as the atypical combination of existing knowledge, we introduce an information-theoretic measure of novelty in scholarly publications. This measure quantifies the degree of `surprise' perceived by a language model that represents the word distribution of scholarly discourse. The proposed measure is accompanied by face and construct validity evidence; the former demonstrates correspondence to scientific common sense, and the latter is endorsed through alignment with novelty evaluations from a select panel of domain experts. Additionally, characterized by its interpretability, fine granularity, and accessibility, this measure addresses gaps prevalent in existing methods. We believe this measure holds great potential to benefit editors, stakeholders, and policymakers, and it provides a reliable lens for examining the relationship between novelty and academic dynamics such as creativity, interdisciplinarity, and scientific advances.
\end{abstract}


\section{Introduction}

From alchemy to AlphaFold, the progress of academic endeavor is undeniable.
Much as gene mutation drives evolution, novelty---often seen as the recombination of existing knowledge elements---allows for scientific breakthroughs from known terrain.
It also serves as a key criterion for academic institutions and funding agencies, with submissions frequently required to showcase novel findings and insights.
Of the 1.8 million papers published on PubMed in 2022, more than 8\% included the word `novel' in their titles or abstracts \citep{nih2022numbers}.

While recognizing novelty is crucial for pushing research boundaries, its presence and extent are often subjectively assessed by domain experts.
Given the huge volume of scientific outputs and applications, an automatic measure of scholarly novelty can be beneficial for reviewers, editors, stakeholders, and policymakers.
The pursuit of such a measure, in alignment with human judgment, emerges as both a technical and philosophical imperative.

In this study, we conceptualize novelty as `atypical combinations of knowledge' within the framework of information theory, and adopt the information-theoretic measure \emph{surprisal} to evaluate the novelty present in scholarly publications based on their \emph{content} in the context of current scholarly landscapes.
Specifically, we treat each word in the main content of a manuscript as a unit of knowledge.
The atypicality of a sequence of these units, i.e., a manuscript, can be calculated based on the sequence's deviation from a distribution that approximates word distribution in scholarly discourse.
This enables the identification and measurement of novelty in a bottom-up manner, from the word level onward.
Furthermore, the measure's numeric values are interpretable due to its information-theoretic nature, a characteristic not possessed by other existing novelty measures.
The measure, code, and data are accessible under a permissive license, available at \href{https://github.com/Wang-Haining/noveval}{github.com/Wang-Haining/noveval}.

\section{Literature Review}\label{sec: litrev}

Historical inquiries have shed light on the defining characteristic of novelty in scholarly publications as the \emph{recombination of existing knowledge in atypical ways}\citep{mayer1995search, simonton2004creativity, bornmann2019we, godart2020sociology}.
We use the terms `originality' and `novelty' interchangeably in the context of scholarly communication \citep{guetzkow2004originality}. 
However, creativity often entails both usefulness \citep{sternberg2011cognitive, pichot2022construct} and impact \citep{fontana2020new, park2023papers}, whereas novelty does not necessarily lead to scientific advances \citep{poincare1910mathematical, godart2020sociology}.

\subsection{Novelty Measures Based on Reference, Keyword, Title, and Abstract}
Operationalizing knowledge components at the levels of journals \citep{uzzi2013atypical, lee2015creativity, wang2017bias, veugelers2019scientific} and documents \citep{trapido2015novelty, matsumoto2021introducing, dahlin2005invention}, the atypicality is obtained by contrasting observed combinations against local or global co-occurrences found in past literature.
Apart from hesitation regarding citations made for non-intellectual \citep{gilbert1977referencing, nisonger2011review} or casual reasons \citep{smith1981citation, callaert2014sources, nagaoka2015use}, the main critique is that the granularity of the knowledge unit is coarse and ignores the content of individual manuscripts.
With co-occurrence networks, knowledge units are also operationalized using keywords, where rarely paired keywords \citep{tahamtan2018creativity} and controlled vocabulary \citep{boudreau2016looking} are used as heuristics to gauge novelty.
Semantically divergent titles \citep{jeon2023measuring} and abstracts \citep{shibayama2021measuring} are also considered as indicators of novelty.

\subsection{Novelty Measures Based on Content}
Interestingly, the content of a manuscript is rarely examined.
A straightforward way to explore the full text of a manuscript is to cast novelty evaluation as a regression task \citep{kang2018dataset, wang4360535measuring} using discriminative models, perhaps enhanced with reviews from experts \citep{kang2018dataset} and social media \citep{li2022identifying}.
However, discriminative models assume that the word distribution in the papers being examined has an independent and identically distributed (IID) relationship with respect to the training data.
Therefore, it is theoretically untenable for such models to be applied to papers that do not conform to the IID assumption, as this violates the definition of novelty, which entails unconventional distributions of words.
Using the concept of terminology life stages and distance in the semantic space of contextual embeddings, \citet{luo2022combination} examine rare combinations of terms in the question and method sections of a paper, while possibly overlooking other sections where novelty may also be present.

\subsection{Considerations for Designing a Novelty Measure}
We distill the essential attributes of an effective novelty measure by drawing upon the identified limitations of current measures and incorporating the best practices established within relevant fields.
The acronym `CORE' captures the attributes we consider indispensable: a \emph{content}-based novelty measure that is \emph{open}, \emph{reproducible}, and \emph{explainable}.

First, common sense holds that novelty resides in a manuscript's content.
Seeking novelty from the content closely mirrors what a human reviewer would do and clearly delineates between the measuring of novelty and interdisciplinarity \citep{fontana2020new}.
Second, as trust stems from explainability, a numerical value for a measure cannot be relied upon unless it can be adequately interpreted.
Explainability is also closely related to the third consideration, context, which is used for contrasting atypical from commonplace content, thereby distinguishing novelty from known scholarly landscapes.
The evaluation of a word's novelty varies with context; for example, a discussion of graph theory may be perceived differently in a philosophy journal compared to an applied physics setting \citep{runco1993originality, long2014more, acar2017ingredients}.
Fourth, simply stretching network theories or examining semantic distance does not provide sufficient granularity when investigating whether a snippet of a paper is novel or less so.
A smaller, relevant knowledge unit can enable a more thorough examination for novelty.
Lastly, most novelty measures have heavily relied on proprietary, paywalled citation databases. 
While these databases are useful, the reality is that the vested interests of dataset publishers, such as subscriptions, promotions, and language services, present challenges to the principles of accessibility and transparency in academic research.
An ideal solution would involve releasing measures under permissive licenses, accompanied by all the essential materials required for replication, while avoiding alignment with commercial interests.

\section{An Information-Theoretical Understanding of Novelty in Scholarly Publications}\label{sec: novelty_understanding}
Let us further ground the defining characteristic of novelty, `atypical recombination of existing knowledge units,' in the language of probability and information theory.

First, because a scholarly manuscript includes a sequence of words as its primary content, we consider the unit of knowledge to be words, composed of letters, punctuation marks, numerals, and special characters.
Second, because the narrative is naturally sequential, `recombination' means a sequence of words.
Third, `atypical' essentially means that the likelihood of observing a particular event is low.
Put together, scholarly novelty in an article implies that \emph{the joint probability of observing the sequence of words is low} in the universe of scholarly discourse.

In information theory, a low probability indicates a deviation from the expected distribution of a random variable. 
In this context, the random variable represents the occurrence of each word within the realm of scholarly discourse, denoted $P$. 
By measuring a word's probability relative to $P$, we assess its deviation from the expected norm in scholarly discourse.

This deviation is quantified as Shannon information \citep{shannon1948}, or surprisal \citep{tribus1961thermostatics} for short.
The surprisal of a word $x_i$ from a set of possible words $X$, given its probability under the distribution $P$, is computed as its base-2 logarithm:
\begin{equation}\label{eq: surprisal}
    I(x_i) = -\log P(x_i), \quad x_i \in X
\end{equation}
This formula highlights the inverse relationship between probability and surprisal: as the probability of a token decreases, its surprisal increases, indicating a higher level of novelty or unexpectedness in the random variable representing scholarly discourse.

The cumulative surprisal of a sequence of words $x_1, x_2, \ldots, x_n$ in a manuscript, given their joint probability under the distribution $P$, is computed as follows:
\begin{align}\label{eq: cumulative_surprisal}
I(x_1, x_2, \ldots, x_n) = -\log \prod_{i=1}^{n} P(x_i) 
                         = -\sum_{i=1}^{n} \log P(x_i) 
\end{align}


In essence, words that are highly probable in $P$ result in lower cumulative or average surprisal values, indicating their commonality in scholarly discourse.
This scenario mirrors an academic manuscript discussing well-established data or theories. Such a paper's surprisal value is minimal, as it largely reaffirms what the academic community already knows, offering little to no new information.
Conversely, a paper that introduces a novel concept or method significantly deviating from established norms carries a high surprisal value. 
This indicates a substantial divergence from standard scientific expectations and, consequently, a high degree of novelty in the research.

\section{Approximating Scholarly Discourse With a Language Model}\label{sec: approximating_scientific_discourse}

\subsection{Language Modeling}\label{sec: language_modeling}
The novelty of an academic manuscript is gauged by its divergence from the distribution of scholarly discourse $P$.
However, since $P$ is not directly observable, we use a surrogate distribution, $Q$, to approximate typical scholarly discourse.
Specifically, we operationalize distribution $Q$ using a causal language model, where the probability of the presence of a word is conditioned on the preceding words and is estimated from English Wikipedia.

We adopted the Generative Pretrained Transformer (GPT-2) architecture \citep{radford2019language}, employing its smallest model variant with approximately 124 million parameters, and trained it from scratch.\footnote{
Our implementation adhered to the original version, featuring a maximum input length of 1,024 tokens, 12 transformer layers, 12 self-attention heads per layer, and a model dimensionality of 768.
We refrained from applying any dropout or bias terms.}
GPT-2 was selected for its widespread use and the superior ability to model complex language patterns.

For ease of understanding, we narrate using `words' as the unit of knowledge. 
In practice, language models typically use `tokens,' which correspond to shorter words or sub-words, to effectively capture semantic relationships and address out-of-vocabulary issues.
For example, the word `bioinformatics' can be tokenized into `bio,' `inform,' and `atics,' allowing the model to process both familiar and novel terms by breaking them down into recognizable components. 
Token-level granularity renders models sensitive to subtle signals of novelty.
In total, there are 50,304 tokens, the combinations of which constitute academic discourse.

During training, the parameters of the GPT-2 are updated so as to maximize the probability of observing the next token based on its preceding tokens.
If the language model is well-trained, it can serve as a good proxy for scholarly discourse.
For example, given the history `The theory of relativity was proposed by \_\_,' we would expect the language model to predict the next token as `Albert' instead of tokens such as `Ludwig' or `that.'
The probability of observing a sequence from the manuscript of interest is a joint probability of the tokens in the sequence:
\begin{align}\label{eq: joint_probability}
 Q(w_{1:n}) &=  Q(x_1) Q(x_2) Q(x_3)... Q(x_n)\nonumber \\
            &=  Q(x_1) Q(x_2|x_1) Q(x_3|x_{1:2}) ... Q(x_n|x_{1:n-1})\nonumber \\
            &= \prod_{i=1}^{n} Q(x_i|x_{1:i-1})
\end{align}
Note that, in theory, each token's probability is conditioned on \emph{all} its preceding tokens (or history). 
The joint probability can then be readily plugged into Eq.~\ref{eq: cumulative_surprisal} for cumulative surprisal calculation.

\subsection{Modeling Scholarly Discourse}\label{sec: modeling_science_or_not}
The English Wikipedia corpus was created by cleaning a publicly available dump of all Wikipedia articles as of March 2022 \citep{wikimedia}.
The training set contained approximately 6.5 million documents and 4.6 billion tokens.\footnote{We used four A100 GPUs (40 GB memory each), set a batch size of 16 per GPU, and accumulated 5 gradients before updating the model's parameters. This results in a total of approximately 0.33 million tokens per update. We fixed a total training course of 141,000 steps, resulting in a total of approximately 46 billion tokens processed (ten times the size of the Wikipedia corpus). AdamW optimization was used with $\beta_1 = 0.9$ and $\beta_2 = 0.95$. The optimizer had a weight decay rate of $0.1$, and the learning rate was reduced from a maximum of $6\text{e--}4$ to a minimum of $6\text{e--}5$ over the training course, with a warm-up period of 2,000 steps.
The training of the GPT-2 took approximately 43 hours.}

Wikipedia articles represent one genre of scholarly discourse that offers a wide range of facts, common sense, and established theory in diverse research fields \citep{patton2002role, petroni2023improving}.
They share the explanatory nature of typical academic manuscripts. 
Additionally, Wikipedia articles are written by many authors spanning a long period, which may help to account for stylistic and temporal variations in language.
Generally, Wikipedia articles are better suited to our goals than corpora that consist only of academic abstracts or full texts parsed from PDF files.
The former fail to provide a more accurate representation of the structure of full-length articles; the latter often contain OCR artifacts and require further consideration of disciplinary balance, which trades off generality and specificity.

\subsection{Notes on Surprisal Calculation}
The accurate estimation of a surprisal score for a token requires a sufficiently long text, as the probability of the token is conditioned on all its preceding tokens (Eq.~\ref{eq: joint_probability}).
In practice, however, constructing a language model to account for an extremely long context window is computationally expensive, so we adopt a fixed window of 1,024 tokens, faithfully mirroring the original GPT-2 architecture.
To achieve reliable estimations, each token's calculation was conditioned on a history of at least 256 or 512 preceding tokens.\footnote{An `end of document' symbol (\verb_<|endoftext|>_) is prepended to the start of each document for these calculations.}

\section{Assessing Face Validity Using Illustrative Examples}\label{sec: face_validity}

Having established the theory of measuring novelty using surprisal derived from a proxy distribution of scholarly discourse, our next step is to assess its face validity. 
To this end, we will start with two token-level examples for illustration in the realm of quantum physics and move to examining surprisal's capacity to distinguish novelty from two groups of sentences paraphrased from the examples.

\subsection{A Token-Level Examination}
\begin{figure}[h]
    \centering
    \includegraphics[scale=0.22]{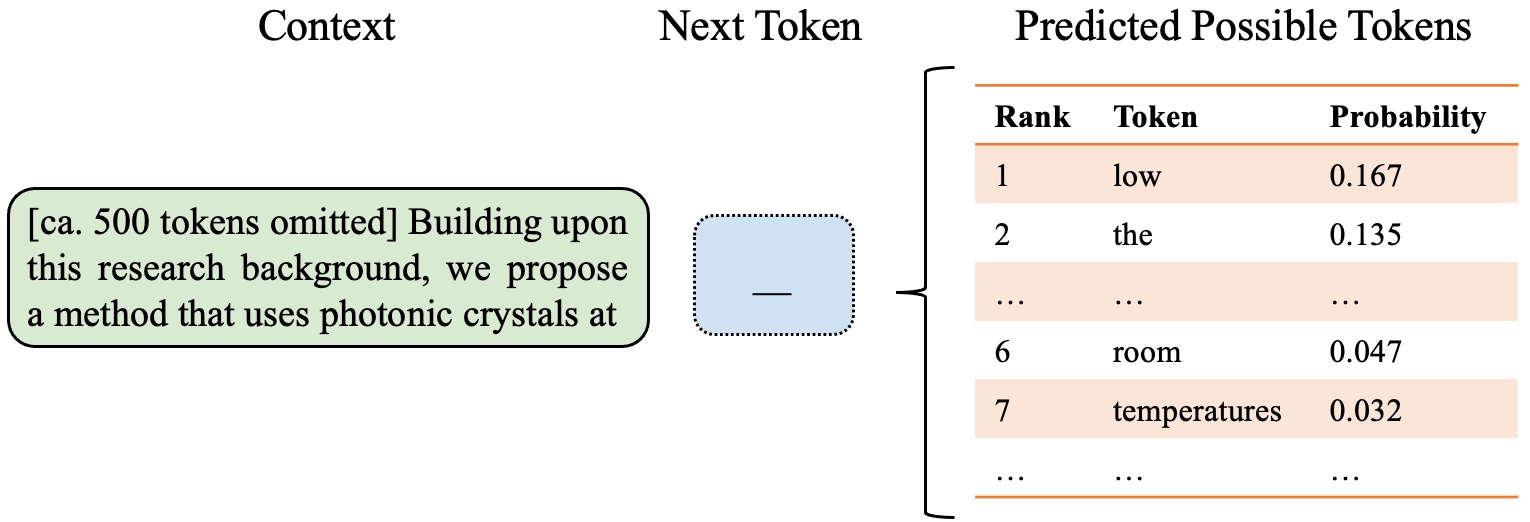}
    \caption{Illustrative examples showing the probability rankings of potential subsequent tokens given a common preceding context. Based on the same history, the GPT-2 model, trained exclusively on English Wikipedia, predicts the next token as `room' with a probability of 0.05 and `low' with a probability of 0.17, indicating that room-temperature observation of quantum states is more novel.}
    \label{fig: token_level_example}
\end{figure}

Consider the following sentence:\footnote{The context text is adopted from Vedral (2014) \citep{vedral2014quantum}.}
\begin{quote}
Quantum physics started with Max Planck's `act of desperation', in which he assumed that energy is quantized in order to explain the intensity profile of the black-body radiation... [ca. 500 tokens omitted] Building upon this research background, we propose a method that uses photonic crystals at \_\_
\end{quote}
An intuitive continuation might be `low temperature,' as current methods typically focus on low-temperature observation.
This is due to quantum states being sensitive to temperature changes, which poses difficulties in maintaining entangled states. 
In contrast, observing quantum entanglement at `room temperature' indicates significant novelty, presenting a formidable challenge to the current capabilities of our technology and equipment.
Although there is only one token difference (i.e., `room' and `low' as shown in Figure~\ref{fig: token_level_example}), we expect the surprisal associated with `room' to be higher than that for `low.'

We calculated the surprisal for the two tokens sharing the same history using the GPT-2 model trained on Wikipedia data (i.e., $Q$).
The token `low' scores a surprisal of 2.58 bits, corresponding to a probability of 0.17 (i.e., $2^{-2.58}$), and is ranked first among all 50,304 tokens in the vocabulary.
The token `room' has a much higher surprisal score of 4.42 bits, corresponding to a lower probability of 0.05 (i.e., $2^{-4.42}$), and is ranked sixth.
This corroborates common sense in quantum entanglement research: performing experiments at low temperatures is commonplace, carrying little novelty, hence higher probability and lower surprisal.

\subsection{A Sentence-Level Examination}
We further tested the face validity at the sentence level using two groups of sentences sharing the same history: each of the two sentences mentioned before was paraphrased twenty times using ChatGPT.
The average surprisal score was calculated for each paraphrase in the two groups.
Our null hypothesis posited that there would be no significant difference in surprisal scores between the groups, with the expectation that the paraphrases of `room-temperature' sentences would not yield higher scores.

A one-tailed Welch's t-test yielded a t-statistic of 2.9 and a p-value of 0.003, with the degrees of freedom approximately 36.5.
This finding indicates a significantly higher average surprisal score for room-temperature paraphrases than for low-temperature ones, thereby reconfirming the effectiveness of surprisal as a measure of novelty.

\section{Assessing Construct Validity Using Known Groups}\label{sec: construct_validity}
Beyond face validation at the token and sentence levels with synthetic examples, we extend our investigation to ascertain the applicability of surprisal at the section level for measuring novelty in a real-world dataset, using the known-groups technique.
This technique is instrumental in testing construct validity, as it involves comparing two distinct groups expected to differ in the construct being measured \citep{portney1993foundations}.
In our case, the validity check may operate on two groups of scientific papers, one of which is perceived as more novel than the other by a group of domain experts.
Creating a dataset consisting of two comparable groups of papers presents significant challenges due to the multifaceted nature of novelty in academic research \citep{guetzkow2004originality}.

\subsection{Authorship Verification Anthology Dataset}
To make the investigation feasible, we focused on novelty in methodology and findings sections and developed the Authorship Verification Anthology (AVA) dataset.
The AVA dataset comprises working notes from a seven-year series of the Authorship Verification task, part of the Conference and Labs of the Evaluation Forum in Digital Text Forensics and Stylometry \citep{pan}.
The shared task annually challenges global teams using shared datasets and requires teams to publish their working notes, detailing their experiments, models, setups, and data processing.
This consistent framework across years ensures a controlled environment for comparing novelty in methodologies and findings sections.

\begin{figure}[htb]
    \centering
    \includegraphics[scale=1]{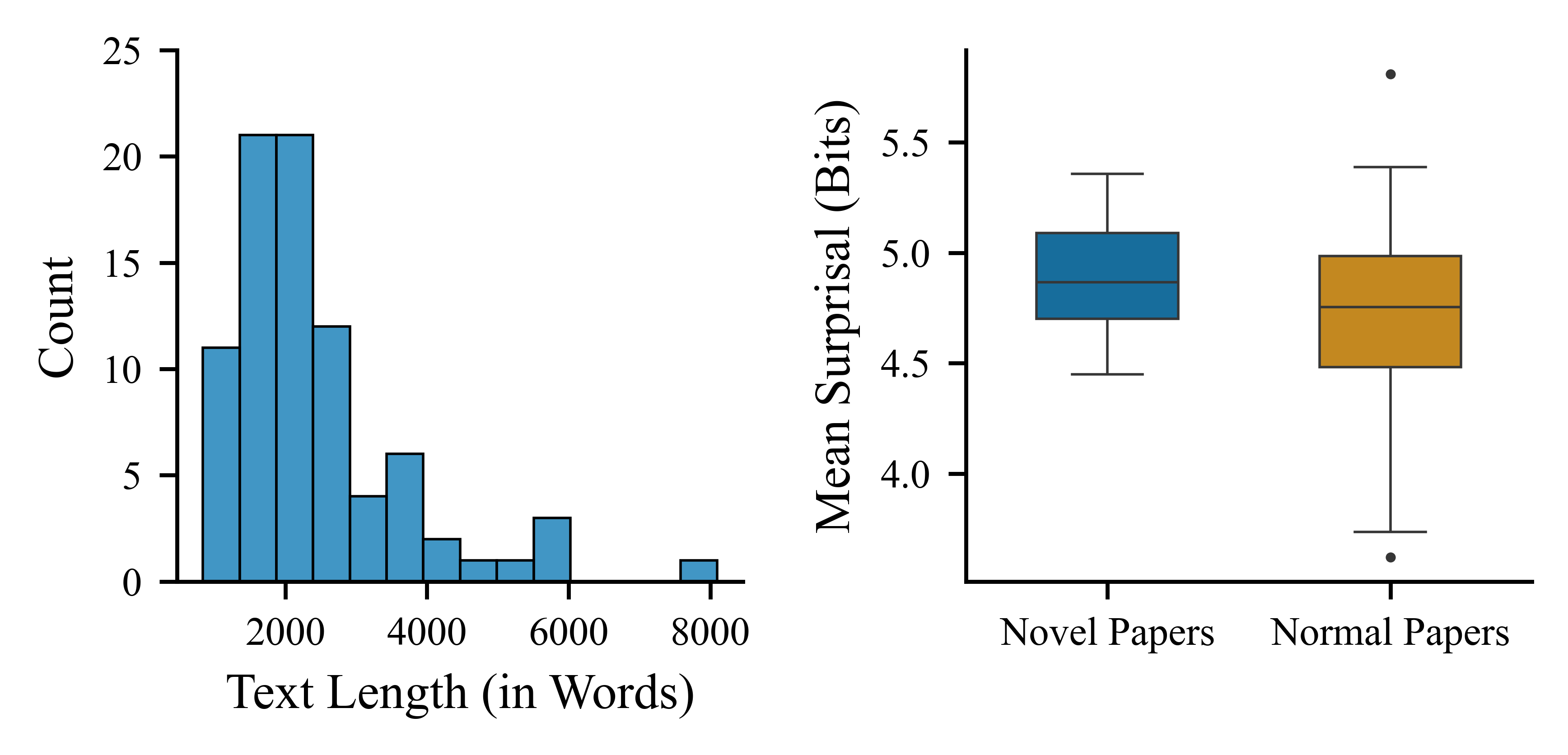}
    \caption{
    The left panel presents the distribution of text lengths in the AVA dataset, measured in words. The right panel shows box plots of the average surprisal scores for two groups of papers, one of which is unanimously deemed more novel according to assessments by domain experts.}
    \label{fig: section_level_example}
\end{figure}

We collected all 83 working notes in PDF format and converted them to plain text. 
Three domain experts independently evaluated each paper using binary scoring to determine the novelty of each working note.
The criteria for novelty focused on methodological innovations in feature engineering, algorithms, and model architecture, as well as novelty from the findings, with a knowledge cutoff in mid-2022.
A conservative threshold was set: papers selected as novel were those unanimously agreed upon by all raters.
The statistics of length distribution and the average surprisal of each manuscript in both groups are shown in Figure~\ref{fig: section_level_example}.

Each paper's novelty is calculated based on the concatenation of abstract and body text using the model $Q$.
We computed the average surprisal of 1,024 tokens, each conditioned on at least 256 preceding tokens, because the papers are typically short (see Fig~\ref{fig: section_level_example}).
Finally, 25 of 80 papers were assigned to the novel group, while the remaining ones formed the normal group.\footnote{Three papers were too short for calculation in our setup.}

\subsection{A Section-Level Examination}

We conducted a one-tailed Welch's t-test on two sets of academic manuscripts categorized as `novel' and `normal,' comparing their average surprisal values derived from a GPT-2 model exposed only to English Wikipedia. 
We found a t-statistic of 2.6 and a p-value of 0.005, with approximately 66.3 degrees of freedom, indicating a statistically significant difference in average surprisal value. 
The novel papers exhibited higher average surprisal than those in the normal group, suggesting greater novelty in their content.

We manually examined the `false alarms'---the eight papers with average surprisal values above 5.0 that were nevertheless deemed less novel by our domain experts---to identify potential causes for these anomalies. 
Two of these papers contain excessive use of in-line mathematical formulas and notations.
One other paper contains nonstandard syntax and informal usage, e.g., directly addressing the audience as `you.'
Three papers, authored by the same research team and reporting a series of incremental improvements on the same method, appear to have their surprisal scores influenced by the consistency in the narrative style of this particular group.
The remaining two papers, with average surprisal scores just above the 5.0 threshold, are considered marginal cases.

\section{Conclusion}

In this proof-of-concept study, we reported a content-based measure for assessing novelty in scholarly publications, accompanied by a Shannonian explanation beginning at the word level.
The measure is reproducible and openly accessible, supported by face and construct validity evidence.
Next, we aim to conduct more rigorous validity tests using datasets involving peers from various global regions \citep{kang2018dataset, bornmann2019we} and in multiple languages.
We will also explore the trade-offs between generality and disciplinary specificity, along with the caveats associated with the use of this measure.

\bibliographystyle{apalike}
\bibliography{main}

\end{document}